\pdfoutput=1
\documentclass[11pt]{article}

\usepackage[preprint]{acl}

\usepackage{times}
\usepackage{latexsym}

\usepackage[T1]{fontenc}
\usepackage[utf8]{inputenc}

\usepackage{microtype}

\usepackage{inconsolata}

\usepackage{graphicx}

\usepackage{svg}
\usepackage{tabularray}
\usepackage{amsmath}
\title{MonoTODia: Translating Monologue Requests to Task-Oriented Dialogues}

	\author{Sebastian Steindl \\
	Ostbayerische Technische \\ Hochschule Amberg-Weiden \\ Germany \\ s.steindl@oth-aw.de \\\And
	Ulrich Schäfer \\
	Ostbayerische Technische \\ Hochschule Amberg-Weiden \\ Germany \\ u.schaefer@oth-aw.de \\\And
	Bernd Ludwig \\
	University Regensburg \\ Germany \\
	bernd.ludwig@ur.de
}

\begin{document}
\maketitle

\begin{abstract}
Data scarcity is one of the main problems when it comes to real-world applications of transformer-based models.
This is especially evident for task-oriented dialogue (TOD) systems, which require specialized datasets, that are usually not readily available. This can hinder companies from adding TOD systems to their services.
This study therefore investigates a novel approach to sourcing annotated dialogues from existing German monologue material.
Focusing on a real-world example, we investigate whether these monologues can be transformed into dialogue formats suitable for training TOD systems.
We show the approach with the concrete example of a company specializing in travel bookings via e-mail. 
We fine-tune state-of-the-art Large Language Models for the task of rewriting e-mails as dialogues and annotating them.
To ensure the quality and validity of the generated data, we employ crowd workers to evaluate the dialogues across multiple criteria and to provide gold-standard annotations for the test dataset.
We further evaluate the usefulness of the dialogues for training TOD systems.
Our evaluation shows that the dialogues and annotations are of high quality and can serve as a valuable starting point for training TOD systems.
Finally, we make the annotated dataset publicly available to foster future research\footnote{\url{https://github.com/sebastian-steindl/MonoTODia}}.
\end{abstract}

\section{Introduction}
\begin{figure}[t]
    \centering
    \includegraphics[width=\linewidth]{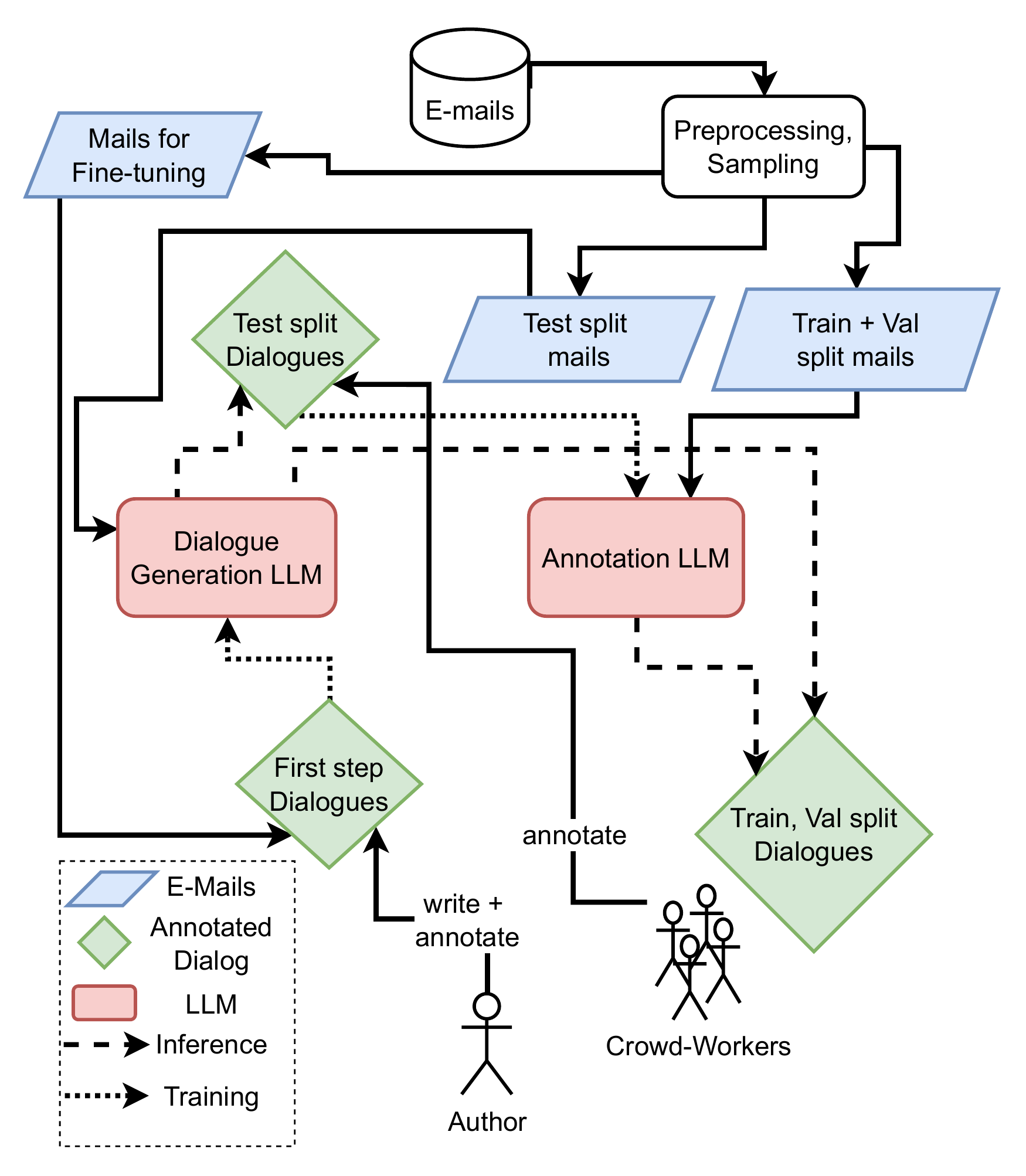}
    \caption{The MonoTODia approach. Blue marks e-mails, green annotated dialogues, and red LLMs. Dashed arrows mark inference, dotted arrows training.}
    \label{fig:mainFig}
\end{figure}
The rise of Large Language Models (LLMs) has inspired many new fields of research and applications.
One of the factors enabling their success is their capability to follow natural language prompts \cite{zhangInstructionTuningLarge2023}, increasing and simplifying control over the model's output.

In general, chatbots can be roughly categorized into Task-Oriented Dialogue (TOD) systems and Open-Domain Dialogue systems \cite{niRecentAdvancesDeep2023}.
%This classification is sometimes extended by Question Answering systems \cite{deriuSurveyEvaluationMethods2021}.
TOD systems can be seen as a natural language interface to one (or multiple) external services, helping the users to achieve a certain task. 
These external services can often be treated as a database or an endpoint that is being queried. The request will be constructed in predefined slots that are being filled by the TOD system during the conversation.
Everyday examples include actions like booking a restaurant or a train ticket.
Furthermore, multiple domains can be combined within one dialogue, enabling the user to, e.g., book a complete vacation, including flights, hotel, and restaurants, within one conversation.
For real-world productive use, the requests to the services will usually need to be made on live data, e.g., to get current prices and availabilities.
The TOD system has to complete three subtasks to fill slots and build requests to the external services: understanding the user (natural language understanding, NLU), deciding on how to react (policy planning, PP), and finally creating a response (natural language generation, NLG) \cite{heGalaxyGenerativePretrained2022}.
Compared to open-domain dialogues, TODs are usually multi-turn but short, constrained to certain domains, and highly structured  \cite{deriuSurveyEvaluationMethods2021}.
While TOD systems were traditionally rule-based, current approaches use deep learning and transformers \cite{su-etal-2022-multi, bang-etal-2023-task, zhaoDescriptionDrivenTaskOrientedDialog2022}, achieving better results but requiring large amounts of training data.
%Additionally, they combine all three subtasks (NLU, PP, and NLG) into one end-to-end model instead of having one model per subtask.

\begin{figure*}
    \centering
    \includegraphics[width=\linewidth]{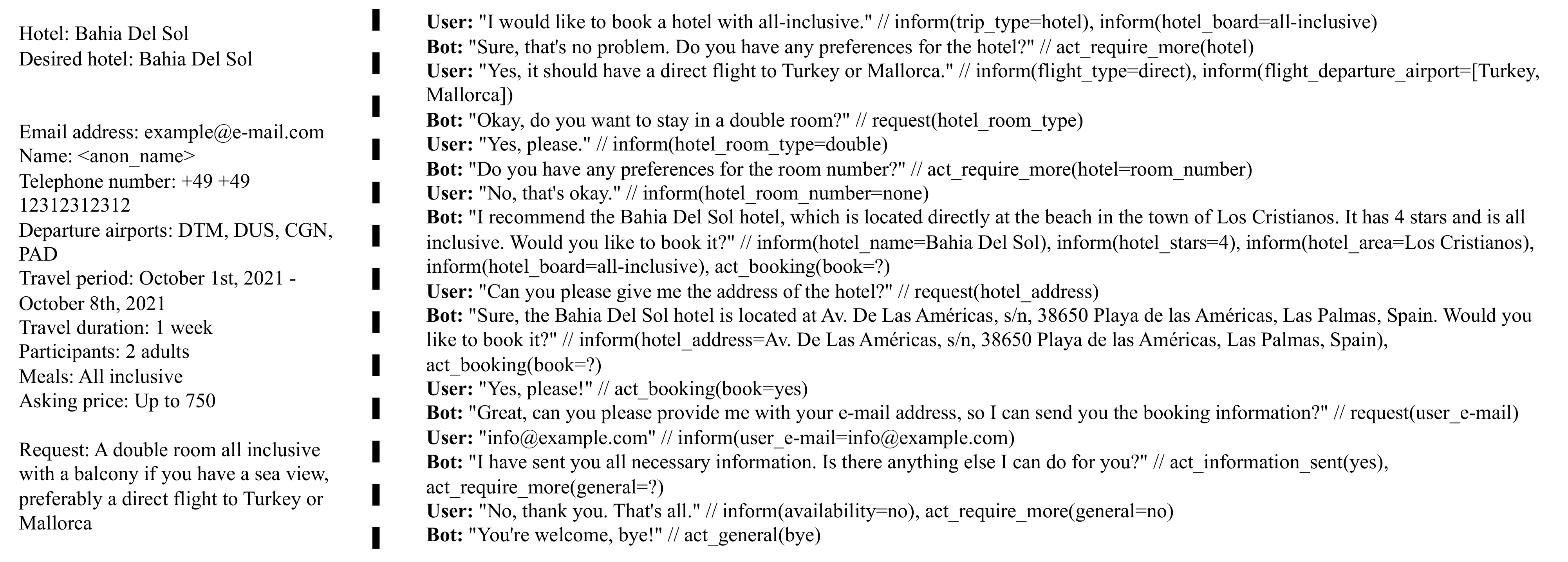}
    \caption{An example e-mail from the corpus after pre-processing on the left and the resulting annotated dialogue after applying the MonoTODia approach on the right.}
    \label{fig:exampleMailDialogue}
\end{figure*}

This work thus studies whether the current advances in LLMs can make training a TOD system more accessible by translating existing monologue data into annotated, task-oriented dialogues.
We fine-tune a state-of-the-art LLM to automatically translate the monologues into dialogues. In a second step, they are annotated with a LLM\@.
The method is demonstrated on real-world e-mail requests.
To assess the quality of the resulting dialogues and annotations, we perform human evaluation and investigate the usefulness of the data for training of downstream TOD systems.
Our results indicate that style translation with LLMs could be a viable approach to cold start and low-resource problems for TOD systems.
We publish the resulting dataset with gold-standard annotations for the test split.
An example of an e-mail and the resulting dialogue is shown in Fig.~\ref{fig:exampleMailDialogue}.
Further examples of dialogues generated with MonoTODia are shown in Figures \ref{fig:ExtraExample1} and \ref{fig:ExtraExample2} in Appendix \ref{appendix:MoreDias}.

\section{Problem statement}

The need for training data is aggravated by the special requirements for the data in TOD systems, making data collection tedious, expensive and thus a fundamental bottleneck for the development of TOD systems \cite{axmanContextualDataAugmentation, kulkarni2024synthdst, liControllableDialogueSimulation2022}.
In collaboration with a German enterprise, we thus investigate an approach to tackle this problem: translating existing non-dialogue data to multi-turn TODs. We showcase this on e-mails, which can be seen as monologue requests in this scenario.
This would drastically reduce the data collection and labeling effort while staying close to real-world, domain-specific data and tackle the cold-start problem of dialogue systems.
For the company, such a system would greatly improve their service portfolio.
We treat this question on an exemplar dataset derived from a German SME\@.
The higher-level goal of this company is to digitalize and automate travel bookings.
They collaborate with travel agencies, where they receive travel requests by e-mail and respond with a list of recommendations.
These e-mails contain diverse, often unstructured pieces of information in various amounts and levels of detail, increasing the complexity of translation immensely. 

A dialogue system is well-suited for booking scenarios since it allows for filling the needed slots and offers the possibility to, \emph{bidirectionally}, ask for additional information, make proposals, and change previous slots.
This interactivity mimics the interaction between a user and a respective human counterpart much more closely than an e-mail. Moreover, such a system would speed up processes because the response time in synchronous communication channels (chats) would be generally shorter than for asynchronous commonuication (e-mails).

The goal of the intended TOD system would be to assess the user's needs and wants. The final, legally binding confirmation of the booking would happen through a second communication channel. 
%For this last step, the user needs to provide critical personal information, such as bank details.
%While in theory a TOD chatbot could be designed to also handle this, it is a design choice to keep it separate because giving this type of information to a chatbot might cause suspicion or mistrust among clients.

Training a TOD system on e-mails directly is not possible since, e.g., the format and style don't fit, they lack the chatbot speaker role, and they are not annotated.
The translation from e-mails to dialogues is complex and infeasible with traditional algorithms for multiple reasons.
Firstly, one has to be able to identify all domain-specific relevant information within the e-mail, which is a NLU task.
Then, one has to generate user and system utterances, which entails all the conundrums of NLG\@.
Moreover, this NLG will in many cases need to include new, contextually relevant information that was not given in the e-mail.
For example, if the e-mail only contains the destination, the chatbot would have to ask for, e.g., the travel period. Therefore, some information will need to be invented, i.e., \textit{hallucinated}.

The recent advances in LLMs could offer an elegant solution to all of these tasks, with guided hallucinations even being desired to some extent.
To enable this project, we defined a specific ontology (cf. Tab.~\ref{tbl:ontology} in the Appendix \ref{appendix:Ontology}) will be the basis for the LLM prompts and dialogue annotation.
To the best of our knowledge, there is no current method that could be used for the problem that we tackle.

\subsection{Existing Data and Pre-processing}
\begin{figure}[bh]
	\centering
	\includegraphics[width=\linewidth]{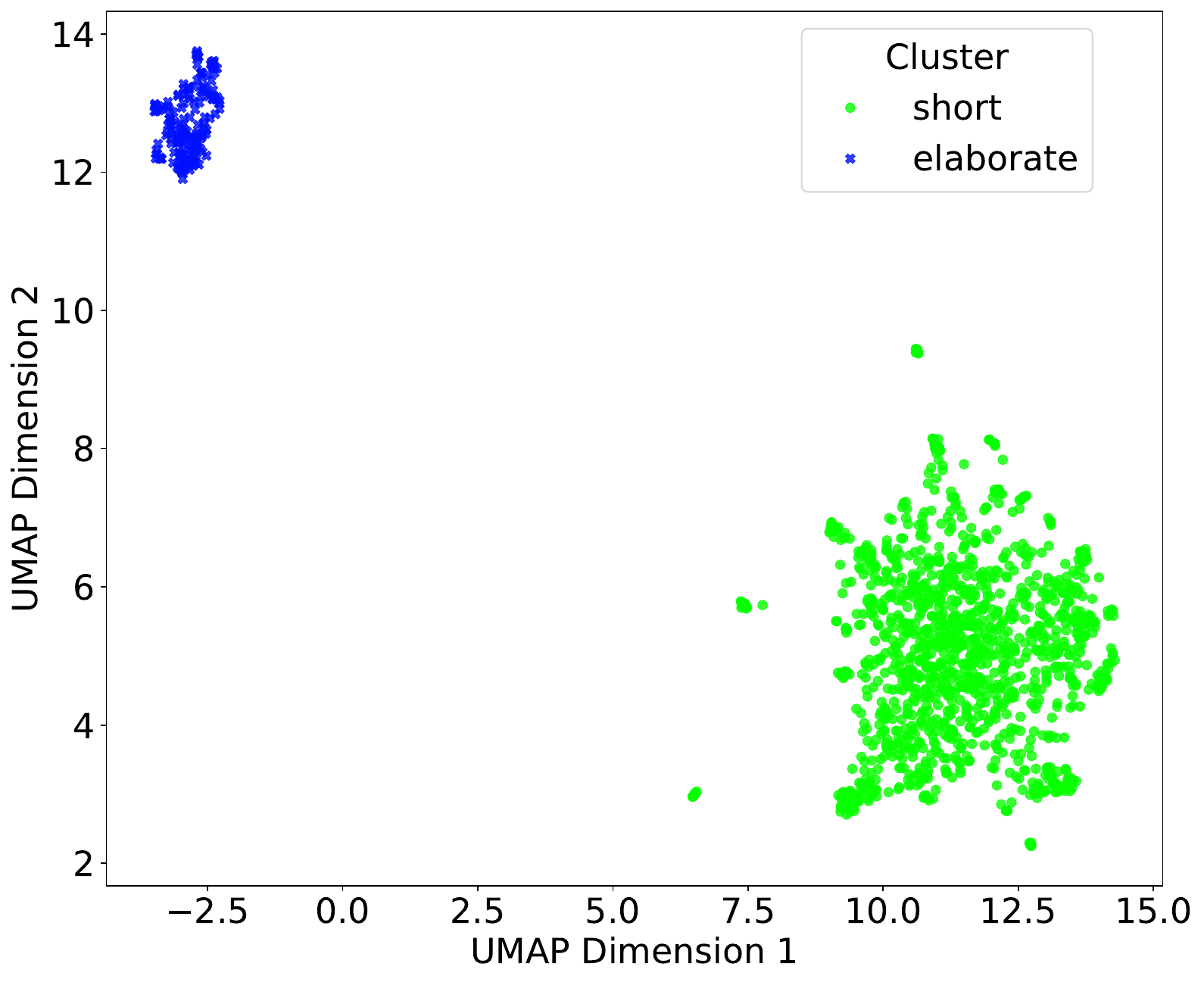}
	\caption{The clustering of the e-mails used for the train split. We first convert the e-mails with TF-IDF and then encode them with UMAP to build the clusters. It is clear that the short e-mails are the majority.}
	\label{fig:clustering}
\end{figure}
The existing data comes from an uncurated database dump of e-mail requests with highly heterogeneous styles.
They range from minimal one-line e-mails, e.g., ``Namibia individual trip'', to elaborate, prose-like free texts and e-mails that give detailed information in an enumeration style.
We can roughly cluster the e-mails into either short or elaborate, as is shown with in Fig.~\ref{fig:clustering}.

Since the data is raw, it includes a significant amount of noise. For example, out-of-office notifications, empty e-mails, test messages, and even apparent scam attempts.
We apply a rigorous rule-based filtering to exclude noise.
This affected roughly 10\% of the full dataset.
%The crowd-worker evaluation later on revealed an additional 10\% that was deemed invalid for the task at hand (cf. Section~\ref{sec:Eval_DiaGen}).
%Therefore, we can assume that roughly 20\% of the e-mails are either noise or not ideal for being translated to e-mails.
Moreover, we anonymized the data to remove personal client information.
% This is not only important for the publication of the data but also for our evaluation, during which crowd workers rate the dialogues based on the e-mail.
The e-mails are nearly exclusively in German. However, our preliminary experiments showed that the used LLM has poor performance on German text. We therefore applied one further step of pre-processing, translating the e-mails from German to English with the Google Translate API\footnote{\url{https://github.com/ssut/py-googletrans}}.
Finally, we construct the train, validation and test datasets by randomly sampling 1500, 150, and 200 e-mails, respectively, ensuring each e-mail is part of only one split.
In summary, our data preparation consists of filtering, anonymization, translation and sampling for the data splits.

%-----------------------------------------------------------------------------------%

\subsection{Impact on Real-World Business Problems}
The travel-booking domain has a high potential for automation. Whereas online booking is nowadays established as an alternative to travel agencies, these services mostly rely on the user filling out static forms.

The cooperating company, adigi GmbH\footnote{\url{https://www.adigi.ai}}, is working towards interactive, natural language travel-booking, offering cloud-based B2B solutions.
Compared to manual request processing, this leads to increased speed and reduced cost. 
Currently, its client base consists predominantly of travel agencies, who act as intermediaries relaying the end-customers' requests via e-mail.
Extending the service portfolio by integrating a TOD system could thus drastically increase the number of clients by opening an additional direct sales channel to the end-customer, promoting business growth and competitiveness.

\section{Background and Related Work}
We will now describe related work for TOD systems, data augmentation and data style translation.

\textbf{TOD systems and datasets.}
TOD systems were traditionally implemented by solving each subtask separately \cite{youngPOMDPBasedStatisticalSpoken2013}.
With the publication of large datasets, the field has moved towards deep learning-based systems such as \citet{lin-etal-2020-mintl, peng-etal-2021-soloist, heGalaxyGenerativePretrained2022}.
Benchmark datasets include, e.g., MultiWOZ \cite{budzianowski-etal-2018-multiwoz},
%This dataset contains a total of roughly ten thousand dialogues, and its publication has massively increased the size of the available benchmark datasets.
KVRET \cite{eric-etal-2017-key} and SGD \cite{rastogiScalableMultiDomainConversational2020}.
% The latter, while larger than MultiWOZ, was, however, not created using the Wizard-of-Oz setup \cite{kelleyIterativeDesignMethodology1984}, but by having crowd workers verbalize utterances that were sampled based on schemata.

% \subsection{Natural Language Interface to Database}
% A TOD system for the present use case will, from a technical point of view, likely have to derive one or multiple database requests from the user's natural language input.
% This task of translating natural language into a database query in a structured language like SQL has been investigated even before the advent of deep learning under the name of \textit{Natural Language Interface to Database} (NLIDB) \cite{nihalaniNaturalLanguageInterface2011,kumarNaturalLanguageInterface2013,liConstructingInteractiveNatural2014}. 
% The desired advantage of this approach was to facilitate the use of database systems and make the interfaces more user-friendly for non-experts, even for complex intents \cite{liConstructingInteractiveNatural2014}.
% However, shortcomings in the NLP part of such a system led to problems in early research \cite{nihalaniNaturalLanguageInterface2011, kumarNaturalLanguageInterface2013}.
% With the general progress in NLP and the availability of public datasets, NLIDB has regained some focus, and progress has been made thanks to deep learning approaches \cite{abbasReviewNLIDBDeep2022}.

\textbf{Data augmentation for dialogues.}
Data augmentation describes the sourcing of synthetic data by applying certain transformations to existing data in order to increase the amount of training data and the model's generalization ability \cite{shortenTextDataAugmentation2021}.
%While, to the best of our knowledge, none of the current state-of-the-art models for the MultiWOZ benchmark dataset actively use data augmentation, different techniques have been studied for TOD systems in general.
Approaches include backtranslation \cite{kulhanekAuGPTAuxiliaryTasks2021}, incorporation of external datasets \cite{xu-etal-2021-caire}, simulating dialogues based on schemata \cite{pengSYNERGYBuildingTask2021}, graphs \cite{grittaConversationGraphData2021}, framing it as a text infilling task \cite{axmanContextualDataAugmentation} or using specially trained generator models \cite{steindl-etal-2023-controlled}.
Nowadays, this line of research has also turned to LLMs.
These methods include, for example, paraphrasing templates, using seed data or adding miscommunications to the dialogues
\cite{liControllableDialogueSimulation2022,kulkarni2024synthdst, chenPLACESPromptingLanguage2023, mehriLADLanguageModels2022, steindl-etal-2025-coprus}.
Recently, \textit{model collapse} \cite{shumailovAIModelsCollapse2024} has been discussed, where a model's performance degrades with every iteration of it being trained on model-generated data. 
One way to counteract this is by combining real and synthetic data \cite{gerstgrasserModelCollapseInevitable2024}, which our method does by utilizing the human-written e-mails.

\textbf{Data style translation.}
Translating a text from one ``style'' to another can be interpreted as a special case of NLG and controlled generation.
First, we see summaries, especially abstractive summaries \cite{GUPTA201949}, as one form of such translation.
Furthermore, data-to-text approaches \cite{jagfeldSequencetoSequenceModelsDatatoText2018, sharmaInnovationsNeuralDatatotext2023, wangEvaluatingTextGeneration2021} are relevant applications of this paradigm.
Automatic news writing is another application \cite{Diakopoulos+2019}, as is 
the creation of a dialogue based on a short story \cite{miyazakiDialogueGenerationConditional2023a}.
Further, the HR\-MultiWOZ \cite{xu-etal-2024-hr} dataset is based on schemata that get turned into templates and are paraphrased by an LLM\@.

% With LLMs exhibiting strong NLU and NLG capabilities, one can envision further use cases where the user describes constraints for the generation of natural language being implemented in day-to-day work.
% Examples could include rewriting legal or medical jargon into everyday speech or explaining compiler errors in natural language.

%-----------------------------------------------------------------------------------%
\section{Method}
Our approach uses instruction-tuned LLMs to generate dialogues based on monologue e-mails and subsequently annotate them. 
The LLMs undergo fine-tuning to solve these tasks.
Crowd workers provide gold-standard labels for the test dataset.

The following sections provide a detailed breakdown of the two phases and finally explain the dataset sourcing.

\subsection{Dialogue Generation and Annotation}
We separate the tasks of dialogue generation and annotation into two distinct inference phases, where the model does not see the original e-mail when generating annotations. This prevents information leakage, that could not be reliably stopped with prompt engineering.
When addressing both tasks in a single inference step, the annotation was too informed in many cases. That is, an annotation contained information that could not have been known at this point in the conversation and is only known from the e-mail.
% For example, while the user might only state that he is looking for a hotel in Greece, the annotation 

We argue that this task separation delivers better results due to two reasons. Firstly, it leads to shorter and less complex prompts and task descriptions.
Secondly, if both tasks are done in unison, the model has already attended to the complete information from the e-mail (to generate the dialogue), when annotating the first utterance, provoking information leakage.
Consequently, we create the annotation for every utterance independently of later utterances.
% can be derived from the intuitive assumption that the annotation of the dialogue gets handled utterance-by-utterance.
%The model then hasn't yet ``paid attention'' to the information later in the dialogue. If, on the other hand, the model has seen the e-mail as well, as in the one-step approach, it needs to have paid attention to that information earlier to generate the dialogue. Thus, separating the dialogue generation and annotation might allow the model to disregard information that comes later in the dialogue when annotating an utterance. 
Based on preliminary experiments between various models, we decided to use an instruction-tuned open-source model from the LLaMA 3.1~\cite{llama3} family.
Using an open-source model locally acts as an additional security mechanism, avoiding any risk of uploading client information to an external model provider. 
We use the instruction-tuned model with 8 billion parameters.

To improve the performance of the model, we fine-tuned it utilizing the LoRA \cite{hu2022lora} method for the two tasks separately, resulting in $f_{g}$ for the dialogue generation and $f_{a}$ for the annotation. The details for the fine-tuning are described in the Appendix \ref{appendix:fine-tune}.
For this purpose, we manually created and annotated 20 dialogues $D_{ft} = (x_{ft},y_{ft})$ for e-mails that are not part of any dataset split. This number of dialogues was chosen to allow for some variation of e-mails and dialogues, including different slots and flows, without requiring too much manual labour, since the motivation of our approach is to keep this as low as possible.

%Furthermore, we formulated a prompt for each step that describes the task at hand.
The prompts for each task include an initial description, the task-specific rules, and examples. The examples enable in-context learning, which is known to improve performance \cite{brown2020language}. %stating that the model should translate the given e-mail into a dialogue or annotate the dialogue, depending on the step.
%For the dialogue generation, the instructions contain some requirements that the dialogue should meet and also an example input and output.
To increase output diversity for the dialogue generation, we created three variations of the prompt with different dialogue types examples.
%Since we found that the model often adheres rather closely to the given example, we created three variations of the prompt with the same instructions but different types of dialogue examples to increase diversity.

For the annotation step, we provide the model with general rules for the annotation and all possible slots.
%We also add three examples of annotated dialogues.
The annotation is separated from the utterance in a comment-like style, starting with ``//'' and followed by annotations in the form ``type(slot=value)''. This is the result of preliminary experiments, where this format proved to be more successful and consistent than, e.g., JSON. Furthermore, it is easy to parse. However, other formatting styles are feasible and success might also depend on the specific LLM used. 
The full prompts are shown in Appendix \ref{appendix:FullPrompt}. 

\subsection{Dataset Sourcing}
After pre-processing, sampling, and fine-tuning the dialogue generation LLM $f_{g}$, we generate the dialogues from the e-mails for all splits. We apply light rule-based post-processing, %to prepare them for the evaluation,
% This included \\
mainly removing extraneous tokens before or after the dialogue.

In the second inference phase, we first fine-tuned the annotation LLM $f_{a}^{0}$ on $D_{ft}$ to evaluate the lower bound for the quality of annotations.
We use $f_{a}^{0}$ to predict the annotation for the test set dialogues to compare them to the crowd-worker gold-standard.
Then, we fine-tune $f_{a}^{0}$ additionally on these 200 dialogues with gold-standard annotations, yielding $f_{a}^{1}$.

Notably, the published data uses the gold-standard annotations for the test set and predictions from $f_{a}^{1}$ for the train and validation set. 
%This is because we can use the gold-standard annotations from the crowd workers to fine-tune the annotation model even more and generate better annotations.
%We also evaluate the annotation LLM that was trained on only the 20 original training dialogues, which we wrote and annotated manually, on the crowd-worker annotated test dataset. This serves as a lower bound for the quality of annotations since the amount of training data used to train the final annotation model is a magnitude larger (220 dialogues).
%With this final annotation LLM we predict the annotations for the train and validation dataset and perform light post-processing.
%-----------------------------------------------------------------------------------%
\section{Evaluation}
To evaluate MonoTODia, we evaluate $(i)$ the dialogue generation and annotation in isloation and $(ii)$ the usefulness of the MonoTODia dialogues for training TOD systems.

\subsection{Evaluation of Dialogue Generation}\label{sec:Eval_DiaGen}
\begin{table}[t]
    \centering
    \begin{tabular}{p{1.3cm}p{5cm}}
    \hline
    \textbf{Criteria} & \hspace{1.25cm}\textbf{Short explanation}          \\ \hline
    C-0           & E-mail is a vacation request.                          \\
    C-1           & Information from e-mail is represented in dialogue. \\
    C-2             & User gives more information in dialogue than e-mail.      \\
    C-2-1             & If C-2 is ``Yes'': This additional information makes sense. \\
    C-2-2             & If C-2 is ``Yes'': This additional information is relevant to the booking.   \\
    C-3             & The dialogue follows the rules of creation. \\
    C-4             & The dialogue resembles a real conversation. \\ 
    C-5             & The Bot is helpful to the user.\\ \hline
    \end{tabular}   
    \caption{The criteria and their short explanations for the crowd worker evaluation of the dialogue generation. The exact, full questions are shown in Appendix~\ref{appendix:FullQuestions}.}
    \label{tbl:questions}
\end{table}
    
We evaluate the dialogue generation with the quality of the dialogues per se, and regarding the style translation explicitly.
Both types of evaluation are impossible to perform automatically, since, by definition of the problem, no dialogues exist that allow for reference-based evaluations, ruling out most of the common NLG metrics \cite{gehrmann2023repairing}.
Moreover, multiple aspects of the dialogue quality are intrinsically subjective \cite{amideiUseRatingLikert2019a}.
We therefore opt for human evaluation with crowd workers recruited via Amazon Mechanical Turk to rate the dialogues based on the criteria in Tab.~\ref*{tbl:questions} on a scale of 1 to 5.
These criteria entail qualities such as coherence, relevance, correctness and realness.
For 100 of the test-set dialogues, we collected three independent ratings each.
We ensured the qualification of the raters via a high task approval rate and an additional qualification task. 
They were shown the e-mail, dialogue, and instructions on how the dialogue should be created.
These instructions were derived as closely as possible from the dialogue generation prompt, without giving away that the task was done by a LLM\@.
Moreover, they were given instructions on how to rate the dialogues.

\subsection{Annotation Quality}
To evaluate the annotation generated by the LLM, we opted for a reference-based evaluation by comparing it to crowd workers' annotations for the test data split. 
As such, we used crowd workers to create gold-standard annotations for the test dataset, where its accuracy has the highest importance for the overall evaluation.
%, and also enable a comparison of the LLM annotation to ground-truth.
We ensured crowd-worker qualification as before.% and had every dialogue labeled by one annotator, based on instructions that were as close as possible to the LLM prompt.

\subsection{Complexity of Different E-Mails}
The e-mails that are used as the input of our approach come from various sources and have highly heterogeneous styles.
They range from direct, free-format e-mails to tabular-like information but can be roughly classified as either short or elaborate e-mails (cf. Fig.~\ref{fig:clustering}).
%This depends on the sender, who, in some cases, can be a travel agent.
%These travel agents will then usually follow the same format, which is why we can see that the elaborate cluster in Fig.~\ref{fig:clustering} has a low variance compared to the short cluster.
%
There is no clear indication that either of those two types led to consistently better or worse ratings by the human judges. However, one specific format was over-represented within the worst-rated dialogues. This format presents slot-value pairs (e.g., destination, Europe) separated by multiple line breaks. This very implicit style lacks additional context.
It thus appears that the model struggles to extract the information more than in a more expressive way such as ``slot: value''.

\subsection{Downstream Empirical Evaluation}
While our main focus lies on the translation and its evaluation, we conduct an auxiliary experiment investigating if the generated data is suitable for the training of TOD systems in the Dialogue State Tracking (DST) and response generation (RG) tasks.
To this end, we train two T5~\cite{raffelExploringLimitsTransfer2023} and one BART~\cite{lewis-etal-2020-bart} model. The training details are provided in Appendix \ref{appendix:TOD-Train}.
We formulate the DST task as predicting the dialogue state annotations from the chat history, i.e., all previous utterances.
%While this is similar to the annotation of the dialogues with the LLM, it differs in that the TOD model does not have the description of the annotation and possible slots.
We evaluate this with three metrics: Exact-Match (EM), Soft-Match (SM), and Presence (PR). 
EM measures if there is a perfect match, SM if either all slots or all values are correct, and PR if the ground-truth is a subset of the prediction.
%The EM is 1 if the prediction $\hat{y}$ and ground-truth $y$ are identical sets. 
%The SM is 1 if for each slot $s$ and value $v$ within $y$, either the slot or the value is also in $\hat{y}$.
%The PR measures if $y$ is a subset of $\hat{y}$. 
For each metric, we report the mean percentage over all utterances and dialogues.
Their exact formulations are provided in Appendix \ref{appendix:Metrics}.

For the RG task, we provide the model with oracle annotations and the chat history and evaluate the generated response with the BERTScore \cite{zhang2019bertscore}.
In every of these cases, we use the annotations from $f_{a}^0$ for the train and validation set, and the gold-standard for the test data.
%-----------------------------------------------------------------------------------%
\section{Results and Discussion}
\textbf{Dialogue Generation.}
The outcomes of the crowd worker evaluation is summarized in Tab.~\ref{tab:resultsAMT}, showing the average ratings for each question.
They show that the generated dialogues have high quality, achieving an average rating of at least $4$ out of $5$ in nearly all tested criteria, with the lowest rating being $3.98$.
% The highest average rating of 4.37 was achieved for the helpfulness of the chatbot within the dialogue (C-5).
% The lowest average rating of 3.98 was given for the question of whether all information from the e-mail is represented in the dialogue (C-1).
We see that even after our filtering, the judges deemed 11\% of the e-mails to not be valid input for the task of generating a dialogue. This can mostly be attributed to e-mails being very uninformative (too short), as evidenced by significantly lower scores for C-1 and higher scores for C-2 in the invalid e-mails subset.
When we control for the input to be valid, we can see that every criterion improves. Naturally, the opposite is true when only considering invalid inputs.
%It is also sensible to see that in that case, C-1 is rated remarkably worse, as well as C-2 being a lot higher.
Interestingly, C-4 and C-5 remain on a high level and see only minor changes when controlling for input validity. This underlines the strong language generation skills of the LLM\@. The consistently good scores for C-5 specifically can be attributed to the model being trained with the objective to be a helpful bot itself.
% Regarding the only non-rate question (C-1-1), we see that roughly 34 \% of the dialogues contain more information than the original e-mail.
% This is a desirable result, since many e-mails do not contain all the information one would expect in a booking conversation and would require follow-up e-mails.
% The questions C-1-2 and C-1-3 were only answered if the crowd worker answered \mbox{C-1-1} positively.
% Their average ratings were calculated accordingly.

\begin{table}[]
    \centering
    \begin{tabular}{lccc} 
        \hline
        \textbf{Criteria} & \textbf{Average} & \textbf{Valid} & \textbf{Invalid} \\ 
        \hline
        C-0$^\ast$  (valid)           & $89\%$ & n/a & n/a                 \\
        C-1 (inf. exists)              & $3.98$  & $4.09$ & $2.71$                    \\
        C-2$^\ast$ (more inf.)  & $34\%$ & $30\%$ & $79\%$                    \\
        C-2-1 (sensible)             & $4.27$  & $4.48$ &   $3.37$                \\
        C-2-2 (relevant)             & $4.33$ & $4.50$ &     $3.58$                \\
        C-3 (rules)             & $4.07$  & $4.15$ & $3.17$                  \\
        C-4 (realness)            & $4.41$     & $4.43$ &    $4.25$            \\
        C-5 (helpful)             & $4.37$ & $4.41$ &        $4.00$             \\
        \hline
    \end{tabular}
    \caption{\label{tab:resultsAMT} Average rating for the criteria.  Valid column contains the results for dialogues where a majority of raters judged the input e-mail as valid, i.e., C-0 is positive. Invalid column is analogous. $^\ast$: Binary question, for which we report the percentage of positive answers.}
\end{table}
	
\textbf{Annotation.}
To measure the accuracy of the annotations, we compare the annotations from $f_{a}^0$ to the human annotations for the test set. This provides a lower bound estimate for the annotation quality.
The results are $EM = 25.78$, $SM = 36.77$, and $PR = 43.13$. 
These show that the annotations are not perfect, but surprisingly good for the extremely low amount of training data.
Besides, the annotation of task-oriented dialogues is rather complicated and, in this case, allows for some syntactic variances that are semantically equivalent, e.g., for the format of dates or times.
Even without having a human-generated gold-standard, we can assume that the annotations of the train and validation data split are of higher quality since the model got fine-tuned with the additional 200 human-annotated test dialogues.

% \begin{table}
%     \centering
%     \begin{tabular}{lc} 
%     \hline
%     \textbf{Metric} & \textbf{Result}  \\ 
%     \hline
%     EM              & 25.78         \\
%     SM              & 36.77         \\
%     PR              & 43.13        \\ \hline
%     \end{tabular}
%     \caption{Results of the evaluation of the annotation generation by the LLM trained on 20 examples. \label{tbl:AnnotationEval}    }
%    
% \end{table}

\textbf{Downstream Empirical Evaluation.}
The results for the usage of the MonoTODia data in training TOD systems are presented in Tab.~\ref{tbl:EmpiricEval}.
They show that the MonoTODia dialogues can be a valid starting point for implementing a TOD system and can thus alleviate the cold start problem. 
We see a mostly positive correlation between model size and performance. However, BART-large, even though it is the largest model, performs worse than t5-base on the DST metrics but better in the RQ task.

\begin{table}
    \centering
    \begin{tabular}{lccc} 
    \hline
    \textbf{Metric} & \textbf{t5-base}                  & \textbf{t5-small} & \textbf{BART-large}  \\ 
    \hline
    EM     & 36.38                             & 28.44               & 27.23                \\
    SM      & 60.89                             & 52.33               & 51.29                \\
    PR        & 50.47                             & 37.56               & 35.68                \\
    BERT       & 81.24                     & 79.85               &   85.74             \\
    \hline
    \end{tabular}
    \caption{Results of the DST and response generation evaluation. BERTScore shows the mean of the F1-Score, the standard deviation for all models was $9 < \sigma <10$.
    \label{tbl:EmpiricEval}
    }
\end{table}
%-----------------------------------------------------------------------------------%
\section{Conclusion}
This study investigates the feasibility and efficacy of using LLMs to translate e-mails into annotated task-oriented dialogues for the travel booking domain.
% This includes extending the information present in the e-mail to create complete booking conversations.
By fine-tuning a state-of-the-art open-source LLM, performing extensive human assessment and empirical analysis, we have shown that the generated dialogues are of good quality and suitable for downstream training of TOD systems. Even for input e-mails that lack all necessary information, the dialogues achieved good scores.
% With only 20 dialogues being used to fine-tune the model, the effort remains remarkably low.
Note that the published dataset uses train and validation annotations predicted from fine-tuning on 220 gold-standard dialogues ($f_a^1$) to provide higher-quality annotations. Nevertheless, we observe that even a smaller dataset of only 20 examples can be a sufficient foundation for the training of TOD systems.
The evaluation results show that the generated dialogues closely resemble real conversations, contain relevant information, and that the bot in the conversations is helpful in achieving the user's goal.
Furthermore, the LLM closely followed the rules to generate the dialogue based on the e-mail.
These results are consistent with other studies on synthetic dialogues \cite{mehriLADLanguageModels2022,baeBuildingRoleSpecified2022,chenPLACESPromptingLanguage2023,kulkarni2024synthdst}, even though they follow different paradigms and do not translate from existing data.
Overall, the findings suggest that this approach holds promise for addressing the challenges of data scarcity in training TOD systems.
Even though our study is limited to only e-mails, we think that by leveraging existing data sources, such as e-mails, IT support tickets, or transcribed calls, and employing modern LLMs, companies can thus overcome barriers to deploying TOD systems in their service portfolios.
Moreover, we had to translate the e-mails and proceed with English dialogues, which will need to be translated back into German for the use case in the cooperating company. LLMs that perform better on German are thus of high interest.
We publish the resulting dataset to support future research.

\section{Ethical Considerations}
Widespread ethical usage of AI is an important step towards socially meaningful technological advance and broad acceptance of AI.
Our work shows that LLMs might be used to generate training data for smaller, more specialized models, whose usage is less restrictive.
We believe that synthetic data can to some extent alleviate problems that usually arise during model training, both regarding data scarcity, but also data imbalance. 
This can allow more organizations and companies to use AI in production. 
For the special case of service-agent-like chatbots, that improve a user's experience when using a service, we believe that the possible benefits outweigh the potential risk of, e.g., loss of jobs.
Nonetheless, using LLM generated data will always bear risks of being biased or faulty. 
Furthermore, a dual use might be problematic, when dialogues are being generated to train chatbots with the aim to, e.g., spread fake news or commit fraud.

The payment per task for the human evaluators was calculated to equal an hourly rate of roughly \$10 given the average time needed, exceeding the Federal US minimum wage of \$$7.25$ per hour at the time of writing. Crowd workers were also paid for the qualification tasks.

\section*{Acknowledgements}
We thank the adigi GmbH for their cooperation and making this research possible.

% Bibliography entries for the entire Anthology, followed by custom entries
\bibliography{custom,anthology}
% Custom bibliography entries only
%\bibliography{custom}

\appendix

%\section{Clustering}\label{appendix:Clustering}

\section{Domain-specific Ontology}\label{appendix:Ontology}
\begin{table}[h]
    \centering
    \begin{tabular}{p{1.5cm}p{5.5cm}}\hline
    \textbf{Domain} & \textbf{Slots}                                                                                                                              \\ \hline
    hotel           & board, name, area, address, price, feature, room\_type, room\_amount, stars, transfer, reviews                                              \\
    flight          & departure\_airport, arrival\_airport, airline, type, class, price, duration                                                                 \\
    trip            & travel\_period\_start, travel\_period\_end, length, price, type, destination, guests, guests\_children, availability, confirmation\_number  \\
    user            & name, phone, e-mail                                                                                                                         \\
    act             & require\_more, booking, information\_sent, general         \\ \hline                                                                                
    \end{tabular}
    \caption{The ontology of domains and slots we use as a basis for MonoTODia.}
    \label{tbl:ontology}
    
\end{table}

\section{Fine-tuning Details}\label{appendix:fine-tune}
All training and inference was done on a DGX A-100 320 GB platform, that offers eight 40 GB graphics cards.
We utilized the peft~\cite{peft} library to apply LoRA\@. 
We configured LoRA with the following parameters: 
$r = 16$, $\alpha = 64$, dropout probability $=0.1$, and target the modules q\_proj, up\_proj, o\_proj, k\_proj, down\_proj, gate\_proj and v\_proj.
We do not use a bias in LoRA\@.
We fine-tune with a learing rate of 1e-4 and 3e-5 for four and one epochs, for dialogue generation and annotation, respectively.

\section{TOD Training Details}\label{appendix:TOD-Train}
For the TOD system we train two T5~\cite{raffelExploringLimitsTransfer2023} and one BART~\cite{lewis-etal-2020-bart} model. Their sizes range from 60 million to 400 million parameters. 
We train the models for 10 epochs each with a learning rate of 5e-5 and keep only the best instance based on the validation loss.
We formulate the input and output with special tokens that mark the beginning and end of the chat history, annotation and utterance to generate for the RG task. 
For example, a target output in the DST task might be \texttt{<annot>request:trip\_type</annot>} for the input \texttt{<ctx>User: I am looking for a package deal for our vacation. </ctx>}

\section{Metrics}\label{appendix:Metrics}
	
\begin{equation*}
    EM(y, \hat{y}) = 
    \begin{cases} 
    1 & \text{if } \{y_1, y_2, \ldots, y_n\} \\
    & = \{\hat{y}_1, \hat{y}_2, \ldots, \hat{y}_n\} \\
    0 & \text{otherwise}
    \end{cases}
\end{equation*}

\begin{equation*}
SM(y, \hat{y}) = 
\begin{cases} 
1 & \text{if } \{s_i \mid (s_i, v_i) \in y\} \\
  & \cap \{s_j \mid (s_j, v_j) \in \hat{y}\} \neq \emptyset \\
  & \text{or } \{v_i \mid (s_i, v_i) \in y\} \\
  & \cap \{v_j \mid (s_j, v_j) \in \hat{y}\} \neq \emptyset \\
0 & \text{otherwise}
\end{cases}
\end{equation*}

\begin{equation*}
PR(y, \hat{y}) = 
\begin{cases} 
1 & \text{if } \{y_1, y_2, \ldots, y_n\} \\
& \subseteq \{\hat{y}_1, \hat{y}_2, \ldots, \hat{y}_m\} \\
0 & \text{otherwise}
\end{cases}
\end{equation*}

\section{Full Prompts}\label{appendix:FullPrompt}

\begin{figure*}[bh]
    \centering
    \includegraphics[width=\linewidth]{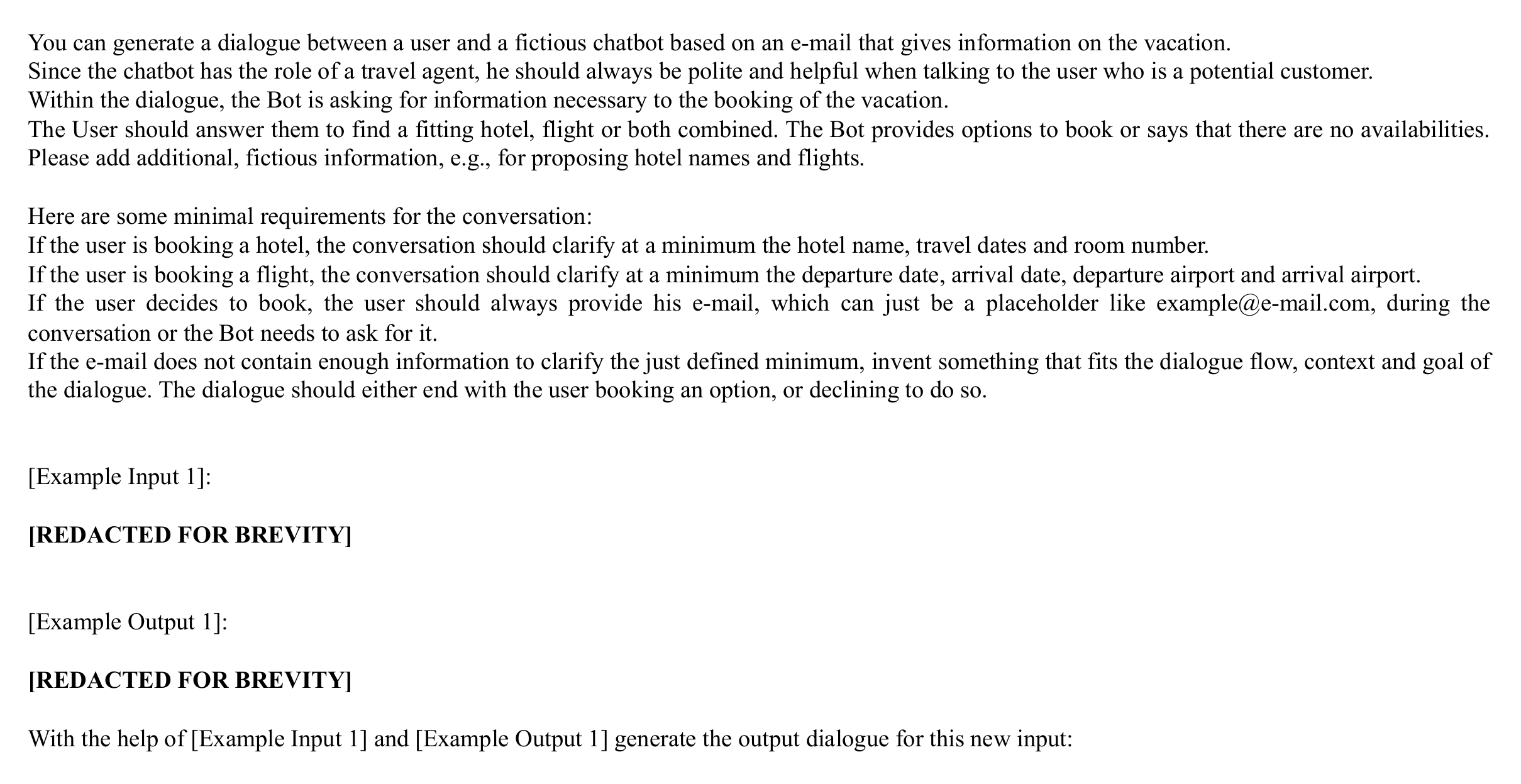}
    \caption{The full prompt used for dialogue generation. Omissions for the sake of brevity are marked in all-caps and bold.}
    \label{fig:fullPromptStep0}
\end{figure*}

\begin{figure*}[bh]
    \centering
    \includegraphics[width=\linewidth]{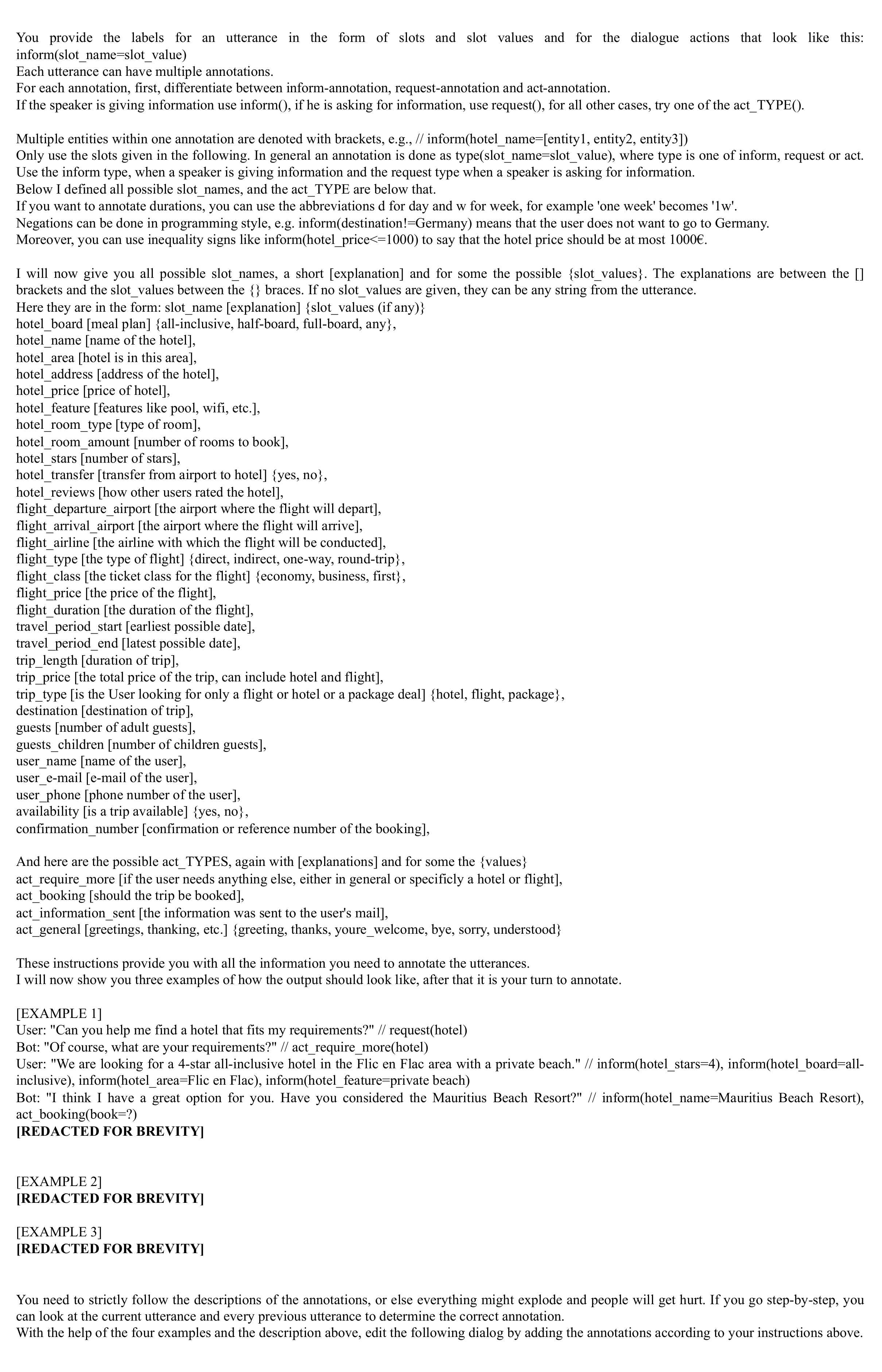}
    \caption{The full prompt used for annotation. Omissions for the sake of brevity are marked in all-caps and bold.}
    \label{fig:fullPromptStep1}
\end{figure*}

\section{Full Dialogue Rating Questions}\label{appendix:FullQuestions}
\begin{table*}[t]
    \centering
    \begin{tabular}{p{2.6cm}p{10cm}}
    \hline
    \textbf{Criteria} & \hspace{2.5cm}\textbf{Full Question}          \\ \hline
    C-0           & Check this box only if the original e-mail is an actual request for vacation offers. Do not check this box, if it is another type of e-mail, such as an empty e-mail, spam or any other e-mail that is not requesting a vacation or information on a vacation.                          \\
    C-1           & On a scale from 1 to 5, how much of the information given in the E-Mail is also represented in the dialog? \\
    C-2             &  Check this box if the user utterances in the dialogue contain more information than was given in the original e-mail.      \\
    C-2-1             & If C-2 is ``Yes'': On a scale from 1 to 5, how much sense does the additional information make in the context of this dialogue? \\
    C-2-2             & If C-2 is ``Yes'': On a scale from 1 to 5, how relevant is the additional information to the booking of a vacation?  \\
    C-3             & On a scale from 1 to 5, how closely does the dialogue follow the instructions for creating the dialogue from the E-Mail (as described above)? \\
    C-4             & On a scale from 1 to 5, how closely does the dialogue resemble a real conversation?  \\ 
    C-5             &  On a scale from 1 to 5, how helpful is the Bot to the User?\\ \hline
    \end{tabular}   
    \caption{The criteria and the full questions as shown to the dialogue raters.}
    \label{tbl:questions_full}
\end{table*}

\section{Further Dialogue Examples}\label{appendix:MoreDias}

\begin{figure*}[bh]
    \centering
    \includegraphics[width=\linewidth]{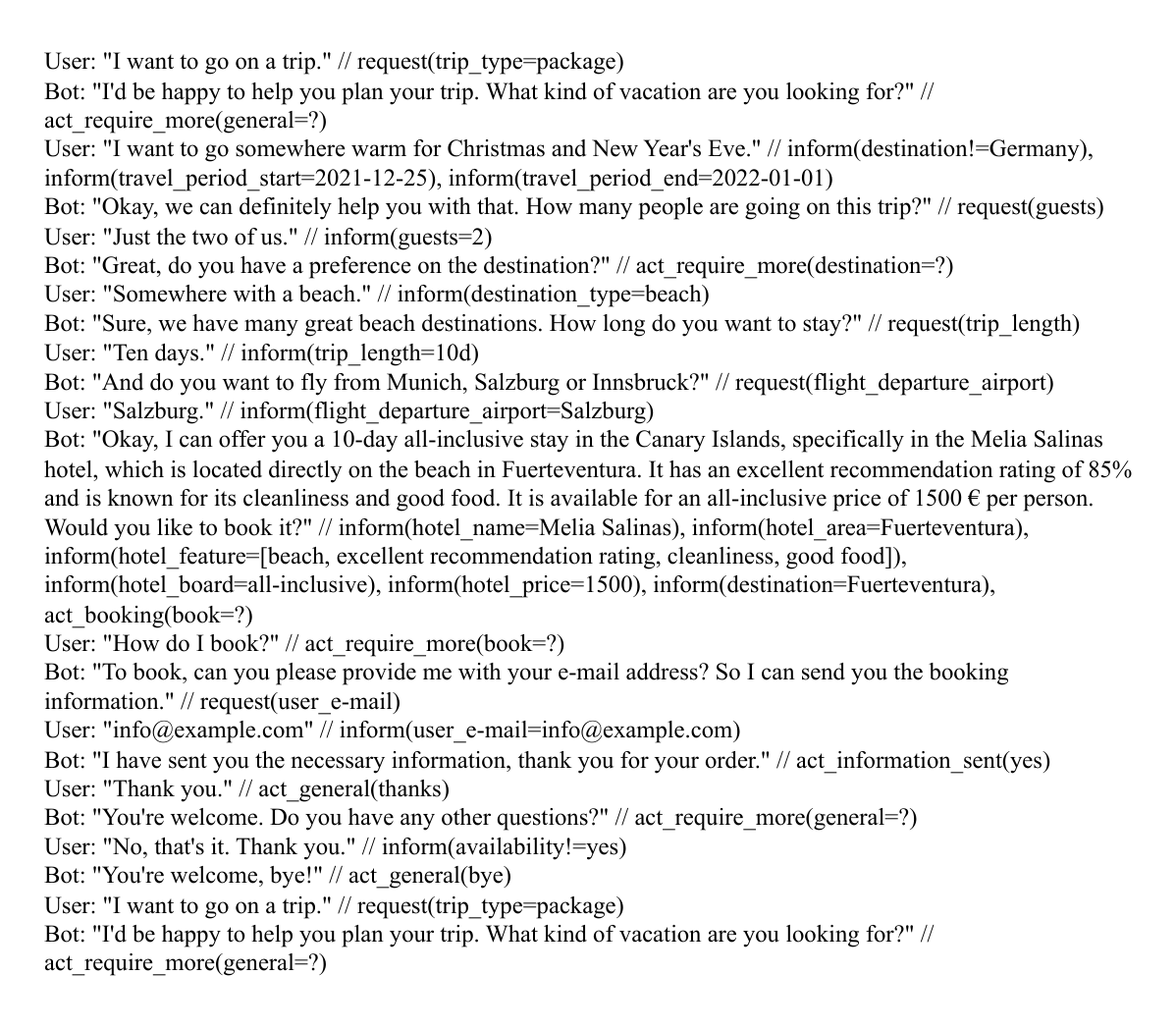}
    \caption{An additional example dialogue generated with MonoTODia.}
    \label{fig:ExtraExample1}
\end{figure*}

\begin{figure*}[bh]
    \centering
    \includegraphics[width=\linewidth]{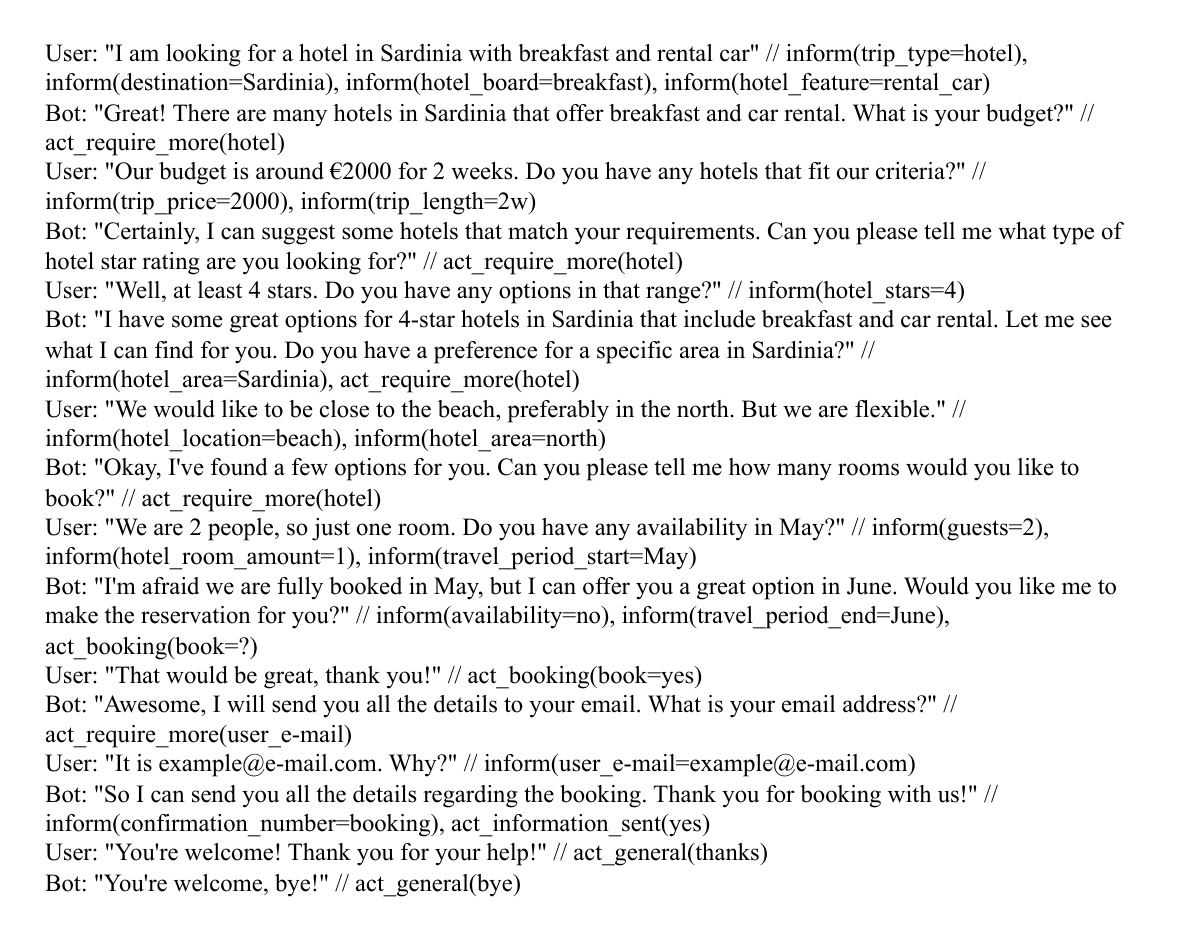}
    \caption{An additional example dialogue generated with MonoTODia.}
    \label{fig:ExtraExample2}
\end{figure*}

\end{document}